\documentclass[11pt]{article}

\usepackage[preprint]{acl}

\usepackage{times}
\usepackage{latexsym}

\usepackage[T1]{fontenc}
\usepackage[utf8]{inputenc}
\usepackage{threeparttable}
\usepackage{microtype}
\usepackage{inconsolata}

\usepackage{graphicx}
\usepackage{caption}
\usepackage{subcaption}
\usepackage{wrapfig}
\usepackage{rotating}

\usepackage{booktabs}
\usepackage{multirow}
\usepackage{makecell}
\usepackage{tabularx}
\usepackage{array}
\usepackage{dcolumn}
\usepackage{bigdelim}
\usepackage{adjustbox}
\usepackage{siunitx}
\usepackage{colortbl}

\usepackage{amsfonts}
\usepackage{amsthm}
\usepackage{mathtools}
\usepackage{nccmath}
\usepackage{bbm}
\usepackage{nicefrac}

\usepackage{algorithm}
\usepackage{algorithmicx}
\usepackage[noend]{algpseudocode}

\usepackage{enumitem}

\usepackage[procnames]{listings}

\usepackage[many]{tcolorbox}

\usepackage{xcolor}
\usepackage{url}

\usepackage{xspace}

\title{O-MARC: Omni Memory-Augmented Compression Distillation for Efficient Video Understanding}

\author{
 \textbf{Peiran Wu\textsuperscript{1}},
 \textbf{Yunze Liu\thanks{Corresponding author.}\textsuperscript{2}},
 \textbf{Chi-Hao Wu\textsuperscript{2}},
 \textbf{Chen Chen\textsuperscript{3}},
 \textbf{Junxiao Shen\textsuperscript{1,2}},
\\
 \textsuperscript{1}University of Bristol,
 \textsuperscript{2}Memories.ai Research,
 \textsuperscript{3}University of Central Florida,
\\
 \small{
   \textbf{Project Web:} \href{https://github.com/WPR001/UGC_VideoCaptioner}{O-MARC}
 }
}

\begin{document}
\maketitle
\begin{abstract}
Omnimodal large language models enable unified audio video understanding, but long joint token sequences make inference costly, and existing benchmarks do not fully isolate audio visual association in noisy user generated videos. We introduce UGC-AVQA, a public UGC benchmark with 1,000 videos and 4,816 QA pairs, where an audio removal test ensures that benchmark questions require both acoustic and visual evidence. To reduce inference cost, we propose OMAC, a training free plug in compression method that preserves salient visual memory and temporally grounded audio anchors. To further make compact models robust to compressed inputs, we introduce O-MARC, a compression distillation framework for learning with memory compressed multimodal contexts. On Qwen2.5-Omni-3B, O-MARC improves the average score across four benchmarks to 45.8, outperforming full token inference at 44.1 and OmniZip at 41.0. OMAC also keeps inference efficient, reducing latency by 34.6\% (1.53$\times$ speedup) and memory by 34.7\% compared with full token inference.
\end{abstract}

\section{Introduction}

Omnimodal large language models have recently emerged as a promising direction for unified audio, visual, and language understanding~\citep{cheng2024videollama,xu2025qwen3,liu2026javisgpt}. Unlike vision-language models that primarily rely on visual inputs~\citep{wang2024qwen2,liu2023visual}, omnimodal models jointly process video frames, audio streams, and textual instructions, enabling reasoning over both what is seen and what is heard. This capability is particularly important for real-world videos, where speech, ambient sounds, scene transitions, and visual actions often provide complementary evidence for understanding complex events. Despite these advances, efficient omnimodal inference remains a significant challenge. Since audio and video are both token-intensive modalities, their joint representation can easily produce long multimodal sequences, leading to substantial memory overhead and slow inference, especially for long videos~\citep{shao2025holitom,shao2025tokens,shen2026fastvid}.

Token compression is a natural solution for reducing this cost, but extending it to audiovisual inputs is non-trivial. Unlike visual-only compression, audiovisual compression must account for two modalities with different redundancy patterns and temporal characteristics. Video tokens often exhibit spatial and temporal redundancy~\citep{wu2025marc,song2024moviechat}, whereas audio tokens may encode brief but decisive cues. Crucially, the importance of each modality is often defined in relation to the other: a sound becomes meaningful only when grounded in the visible scene, and a visual event may remain ambiguous without its corresponding audio. Compressing audio and video in isolation can therefore remove cross-modal evidence that is necessary for accurate reasoning~\citep{tao2025omnizip,ding2026omnisift}.

Another important challenge lies in evaluation. Existing audiovisual benchmarks have covered broad omnimodal understanding, long video reasoning, and temporal alignment, but they do not explicitly isolate joint audio and visual association in noisy user generated videos~\citep{li2025omnivideobench,hong2025worldsense,zhou2025daily}. For example, video understanding benchmarks such as VideoMME~\citep{fu2025video}, TempCompass~\citep{liu2024tempcompass}, and LongVideoBench~\citep{wu2024longvideobench} have been widely adopted to evaluate long video reasoning. Nevertheless, the models commonly assessed on these benchmarks are predominantly vision based and rarely make use of audio signals. To fill this gap, we introduce UGC-AVQA, a benchmark constructed from public UGC short videos. Each question requires evidence from both modalities. To ensure that the benchmark genuinely evaluates audiovisual reasoning, we further apply an audio removal difficulty check: a sample is kept for evaluation only if a strong model fails when the audio track is removed. This procedure filters out questions that can be solved from visual priors alone.

To address this efficiency problem, we propose Omni Memory Augmented Compression, a plug in method inspired by video memory compression, particularly MARC~\citep{wu2025marc}. OMAC keeps compact visual memory and audio anchors, then uses the visual memory distribution to allocate audio budget without modifying the backbone model. Beyond training free inference, we further introduce O-MARC, an omni memory augmented compression distillation framework that teaches omnimodal models to reason under compressed multimodal contexts.
Experiments show that OMAC improves over direct token compression with comparable inference cost across 3B, 7B, and 30B scale models. O-MARC further delivers strong gains in the efficient 3B setting, and when scaled to the 7B model it outperforms the contemporary training based baseline on the overall average. 

Our contributions are threefold. First, we introduce UGC-AVQA, a public UGC benchmark for audio visual association. Second, we propose OMAC as a plug in compression method for audiovisual inputs. Third, we develop O-MARC as a compression distillation framework for robust and efficient omnimodal reasoning.

\section{Related Work}
\label{sec:related}

\subsection{Omni Large Language Model}
In recent years, multimodal systems have undergone continuous development, evolving from large language models to vision language models~\citep{liu2023visual} and, more recently, video large language models~\citep{li2024llava,bai2025qwen3}. However, just using visual information is insufficient for achieving multimodal interaction that closely resembles human perception and communication. This limitation highlights the importance of omni models~\citep{xu2025qwen3,cui2026minicpm}. By incorporating both visual and audio modalities during training and inference, omni models can capture richer cross-modal information, identify fine-grained audiovisual cues that conventional video large language models may overlook, and develop a more comprehensive understanding of inter-modal relationships~\citep{fu2024vita,ge2025arc,li2024baichuan}.
From early omni-modal systems such as GPT-4o~\citep{hurst2024gpt} to recent models in the Qwen-Omni~\citep{xu2025qwen3} and Gemini series~\citep{comanici2025gemini}, it is clear that multimodal research is increasingly moving beyond vision-centric large language models towards omni models that support more integrated audio-visual understanding.

\subsection{Omni Token Compression}
With the development of omni models, these models also encounter the issue of extremely long token sequences when performing audio-visual inference, similar to that faced by large language models for video. Recent research on this issue has focused on token compression to improve the inference efficiency of large multimodal language models~\citep{song2024moviechat,wu2025marc}. This approach is highly effective, as multimodal inputs typically contain a significant amount of redundant information~\citep{bolya2022token,chen2024image,shang2025llava}.
Token compression methods for visual or text data
have been extensively studied, but their application
in audio-visual scenarios has received little attention. Furthermore, we believe that some compression methods such as omnizip ~\citep{tao2025omnizip} must sacrifice some performance in order to improve efficiency, or require large amounts of training data to achieve performance gains such as omnisift~\citep{ding2026omnisift}.

\section{UGC-AVQA Dataset}
\label{sec:dataset}

\begin{figure*}[t!]
    \centering
    \includegraphics[width=0.9\textwidth]{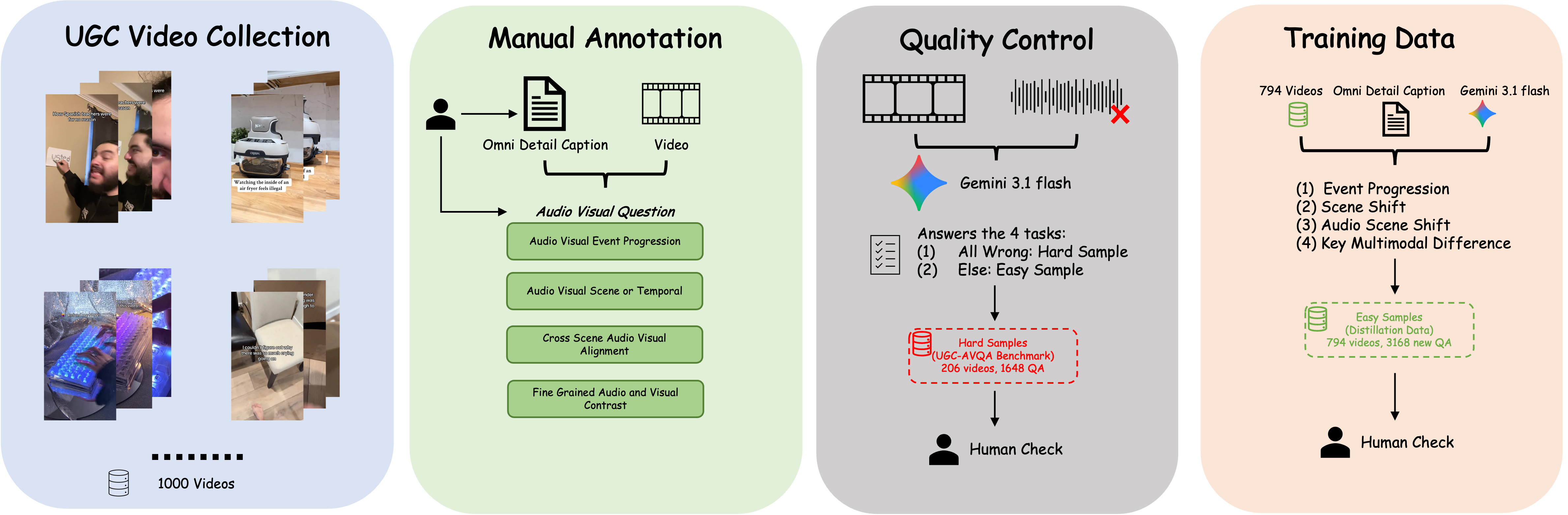}
    \caption{\textbf{UGC-AVQA construction pipeline.} We collect public UGC videos, manually annotate detailed captions and audio visual questions, filter hard benchmark samples with an audio removal test, and review generated QA pairs with trained human annotators.}
    \label{fig:dataset_pipeline}
\end{figure*}

\begin{table}[t!]
    \centering
    \captionsetup{type=table}
    \fontsize{8pt}{9pt}\selectfont
    \setlength{\tabcolsep}{5pt}
    \renewcommand{\arraystretch}{1.12}
    \begin{adjustbox}{width=\columnwidth}
    \begin{tabular}{lccc}
    \toprule
    Dataset & Videos & QA pairs & Focus \\
    \midrule
    WorldSense~\citep{hong2025worldsense} & 1662 & 3172 & Real world omni reasoning \\
    OmniVideoBench~\citep{li2025omnivideobench} & 628 & 1000 & Long video reasoning \\
    Daily-Omni~\citep{zhou2025daily} & 684 & 1197 & Temporal alignment \\
    UGC-AVQA & 1000 & 4816 & UGC audio visual association \\
    \bottomrule
    \end{tabular}
    \end{adjustbox}
    \vspace{-0.2cm}
    \caption{\textbf{Dataset comparison.} UGC-AVQA targets audio visual association in public UGC videos, complementing broader omnimodal and long video benchmarks.}
    \label{tab:dataset_stats}
    \vspace{-0.3cm}
\end{table}

We introduce \textbf{UGC-AVQA}, a user generated video dataset for evaluating fine grained audio visual association in realistic short videos. The dataset is built with human annotated captions and QA pairs, and 30\% of both benchmark candidates and generated training examples are reviewed by trained human annotators and corrected when necessary. Unlike broad video QA benchmarks, UGC-AVQA focuses on questions that require models to connect what is heard with what is seen across events, scenes, and temporal changes. This is especially important for UGC videos, where camera motion, background speech, environmental sounds, and abrupt transitions make compressed omnimodal reasoning difficult.

Table~\ref{tab:dataset_stats} compares UGC-AVQA with recent audio visual benchmarks. WorldSense covers broad real world omnimodal understanding, OmniVideoBench emphasizes long video reasoning, and Daily-Omni includes temporal alignment. UGC-AVQA instead makes audio visual dependence the central target through event progression, scene or temporal transition, cross scene alignment, and fine grained contrast. A detailed category comparison is provided in Appendix~\ref{sec:appendix_taxonomy}.

\subsection{Collection, Annotation, and Compliance}

We collect 1,000 videos from publicly accessible short video platforms and release only the original video URLs, source metadata, detailed captions, and QA annotations, without redistributing raw video files. This preserves provenance and access through the original platforms. The dataset and derived annotations are intended for research on audiovisual reasoning and should be used consistently with the access conditions of the original sources.

Each video is annotated with an omni detail caption covering visual entities, scene transitions, actions, acoustic events, and temporal relations. Human annotators then write multiple choice questions in four categories: \textbf{Audio-Visual Event Progression}, \textbf{Audio-Visual Scene or Temporal Transition}, \textbf{Cross-Scene Audio-Visual Alignment}, and \textbf{Fine-Grained Audio-Visual Contrast}. Representative cases are shown in Appendix~\ref{sec:appendix_examples}.

\subsection{Difficulty-Guided Split Construction}

A central goal of UGC-AVQA is to prevent benchmark questions from being solved by visual priors alone. As shown in Figure~\ref{fig:dataset_pipeline}, we remove the audio track from each candidate sample and evaluate Gemini-3.1-Flash on the visual only input. A sample is included in the benchmark only if the model fails under this audio ablated setting. We then review 30\% of retained candidates with trained human annotators and correct them when necessary.

The resulting benchmark split contains 206 videos and 1,648 QA pairs. The remaining 794 videos form the training split, where Gemini-assisted annotation creates 3,168 QA pairs across four aligned categories: \textbf{Event Progression}, \textbf{Scene Shift}, \textbf{Audio-Scene Match}, and \textbf{Key Multimodal Difference}. We also review 30\% of generated training QA pairs with trained human annotators and correct them when necessary. This yields a difficulty guided curriculum, with easier examples used for training and harder audio dependent examples reserved for evaluation.

\subsection{Evaluation Protocol}

All QA pairs are formatted as multiple choice questions. We report overall and category level accuracy. Since UGC-AVQA is filtered through the audio removal difficulty check, strong performance requires models to preserve acoustic evidence together with visual and temporal context, making it a targeted stress test for audiovisual association under memory compression.

\section{Omni Memory Augmented Compression}
\label{sec:omac}

\begin{figure*}[t!]
    \centering
        \includegraphics[width=\textwidth]{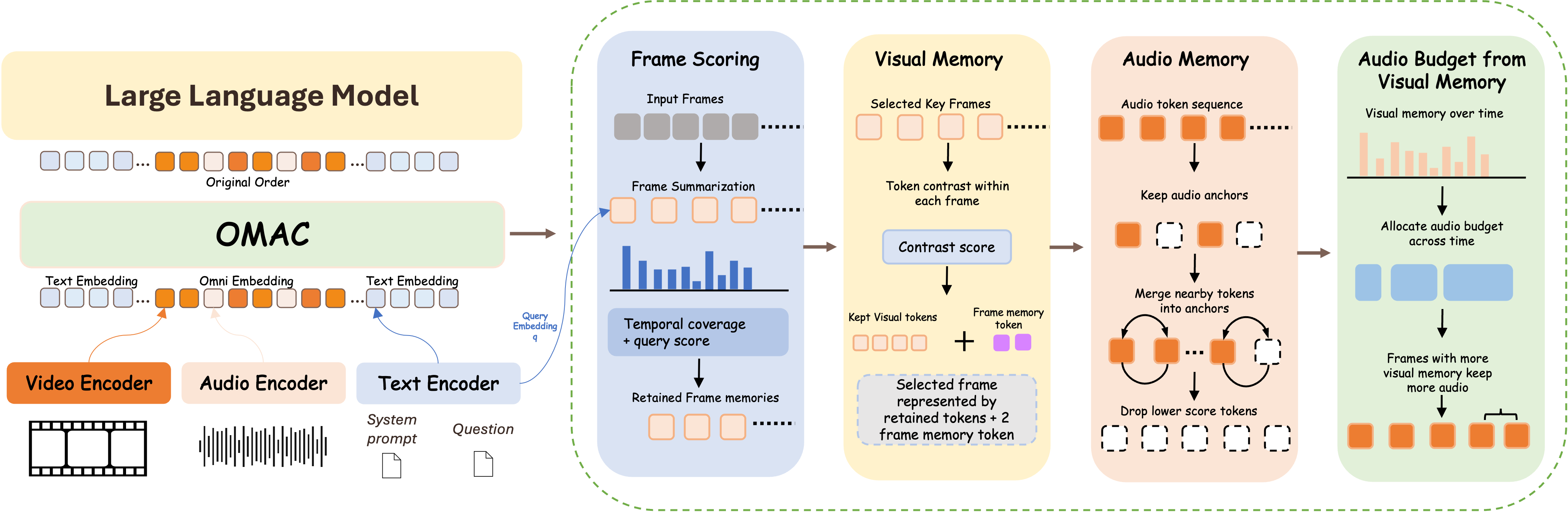}
\caption{\textbf{OMAC for training-free compression.} OMAC keeps informative visual and acoustic cues, forms compact frame memory tokens, and allocates more audio capacity to time regions that receive more visual memory.}
    \label{fig:omac_framework}
\end{figure*}

We propose \textit{Omni Memory Augmented Compression} (OMAC), a method that operates without additional training and compresses long audiovisual inputs by preserving compact memory carriers rather than treating the context as a flat token stream. Let $\mathcal{A}=\{a_i\}_{i=1}^{N_a}$ denote the audio tokens, $\mathcal{V}=\{v_{t,p}\}$ the video tokens, where $t\in\{1,\dots,T\}$ indexes frames and $p\in\{1,\dots,P\}$ indexes visual positions, and $\mathbf{q}\in\mathbb{R}^d$ the query embedding. OMAC outputs a compressed multimodal sequence with audio memory $\mathcal{M}_a$ and visual memory $\mathcal{M}_v$ embedded in the original temporal order. It retains frames that are relevant to the query and keeps audio anchors, forms compact frame memory tokens inside selected frames, and uses the resulting visual memory distribution to adjust the temporal audio budget. The compression strength is controlled by $\rho_a$ for audio and $\rho_v$ for video, while other local selection and merging hyperparameters are fixed across experiments.

\subsection{Frame Scoring \& Visual Memory}

We first summarize each frame into a frame representation
\begin{equation}
\bar{\mathbf{v}}_t=\frac{1}{P}\sum_{p=1}^{P}\mathbf{v}_{t,p},
\end{equation}
where $\mathbf{v}_{t,p}$ is the embedding of the $p$-th token in frame $t$. This frame summary acts as a coarse visual memory descriptor for the whole frame. At this stage, each frame can be viewed as a candidate memory slot for the full frame. We then score each frame by its query relevance:
\begin{equation}
s_t=\cos(\bar{\mathbf{v}}_t,\mathbf{q}).
\end{equation}
Frames with high scores are selected as key frames, and the number of retained frames is controlled by the video compression ratio $\rho_v$. In practice, OMAC uses a selection rule with temporal coverage so that retained frame memories remain both relevant to the query and well spread across the video. In other words, OMAC first chooses which frame memories are worth keeping at high fidelity.

Inside each selected frame, OMAC currently performs contrast filtering within the frame. We compute a token contrast score
\begin{equation}
\alpha_{t,p}=1-\cos(\mathbf{v}_{t,p},\mathbf{c}_t),
\quad
\mathbf{c}_t=\frac{1}{P}\sum_{p=1}^{P}\mathbf{v}_{t,p},
\end{equation}
where $\mathbf{c}_t$ is the centroid of frame $t$. A token receives a high score when it departs from the average content of the frame, which makes it a better carrier of local evidence than repetitive background content. After normalizing $\alpha_{t,p}$, we keep the tokens with the largest scores in the selected frame as explicit visual memory. These retained tokens form the fine grained evidence inside the memory slot for that frame.

We also form a small frame memory token by weighted pooling over the selected tokens:
\begin{equation}
\mathbf{z}_t=
\sum_{p\in\mathcal{K}_t}
\frac{\exp(\hat{\alpha}_{t,p})}{\sum_{j\in\mathcal{K}_t}\exp(\hat{\alpha}_{t,j})}
\mathbf{v}_{t,p},
\end{equation}
where $\mathcal{K}_t$ is the set of retained high score tokens in frame $t$, and $\hat{\alpha}_{t,p}$ is the normalized token contrast score. This compact token serves as an additional frame memory carrier and is inserted back into the original token order using one available token slot in the selected frame. In this way, OMAC preserves both explicit key tokens and a compact summary memory. Therefore, the visual branch has a two level memory structure: frame summaries decide which moments are retained as coarse memory units, and the selected frames are further represented by distinctive local tokens plus a small frame memory token.

\subsection{Audio Memory Budget Adjustment}

For audio, OMAC uses a compression strategy built around anchors. Let $\ell_i$ denote the importance score of audio token $a_i$. We keep high score tokens as audio anchors and merge nearby dropped tokens into them. These anchors are not merely retained tokens; after merging, they become compact acoustic memory slots that summarize nearby evidence. At a coarser scale, OMAC also groups audio tokens into aligned temporal segments, so the audio branch can be viewed as combining fine grained anchor memory with memory allocation over local time regions. If $\mathcal{G}(j)$ is the set of tokens assigned to anchor $a_j$, then the updated anchor feature $\tilde{\mathbf{a}}_j$ is
\begin{equation}
\tilde{\mathbf{a}}_j=
\frac{
\mathbf{a}_j+\sum_{i\in\mathcal{G}(j)}\omega_{ij}\mathbf{a}_i
}{
1+\sum_{i\in\mathcal{G}(j)}\omega_{ij}
},
\end{equation}
where $\mathbf{a}_j$ is the original feature of anchor $a_j$, and $\omega_{ij}$ is the merge weight for token $a_i$. The number of retained audio anchors is primarily controlled by the audio compression ratio $\rho_a$. This gives a compact audio memory while keeping dominant acoustic evidence such as speech cues, sound events, and local temporal context.

OMAC then uses the \emph{resulting visual memory allocation over time} to lightly adjust the temporal audio budget. After visual compression, each selected frame retains a different amount of visual evidence, including explicit kept tokens and a small number of frame memory tokens. We therefore interpret the visual memory retained at frame $t$ as a temporal importance signal. Let $\pi(i)$ map audio token $a_i$ to its aligned frame index, and let $w_t$ be the visual importance weight of frame $t$, derived from the amount of visual memory retained for that frame after visual compression. This induces a coarse temporal partition of the audio stream, where each aligned time segment acts as a candidate temporal memory region. If $B_a$ is the total retained audio budget and $n_t$ is the number of audio tokens aligned to frame $t$, then the target audio budget for frame $t$ is
\begin{equation}
b_t=
B_a \cdot
\frac{n_t w_t}{\sum_{\tau=1}^{T} n_{\tau} w_{\tau}}.
\end{equation}
This step does not increase the total audio budget. It only reallocates more audio capacity to moments that receive more visual memory and compresses visually lighter intervals more aggressively. From the memory perspective, this means OMAC spends more acoustic memory on temporal regions that the visual branch has already identified as worthy of higher memory capacity, which helps the compressed sequence preserve event structure across modalities rather than isolated local tokens.

\section{Omni Memory-Augmented RL Token Compression Distillation}
\label{sec:compression_distill}

\begin{figure*}[t!]
    \centering
        \includegraphics[width=0.9\textwidth]{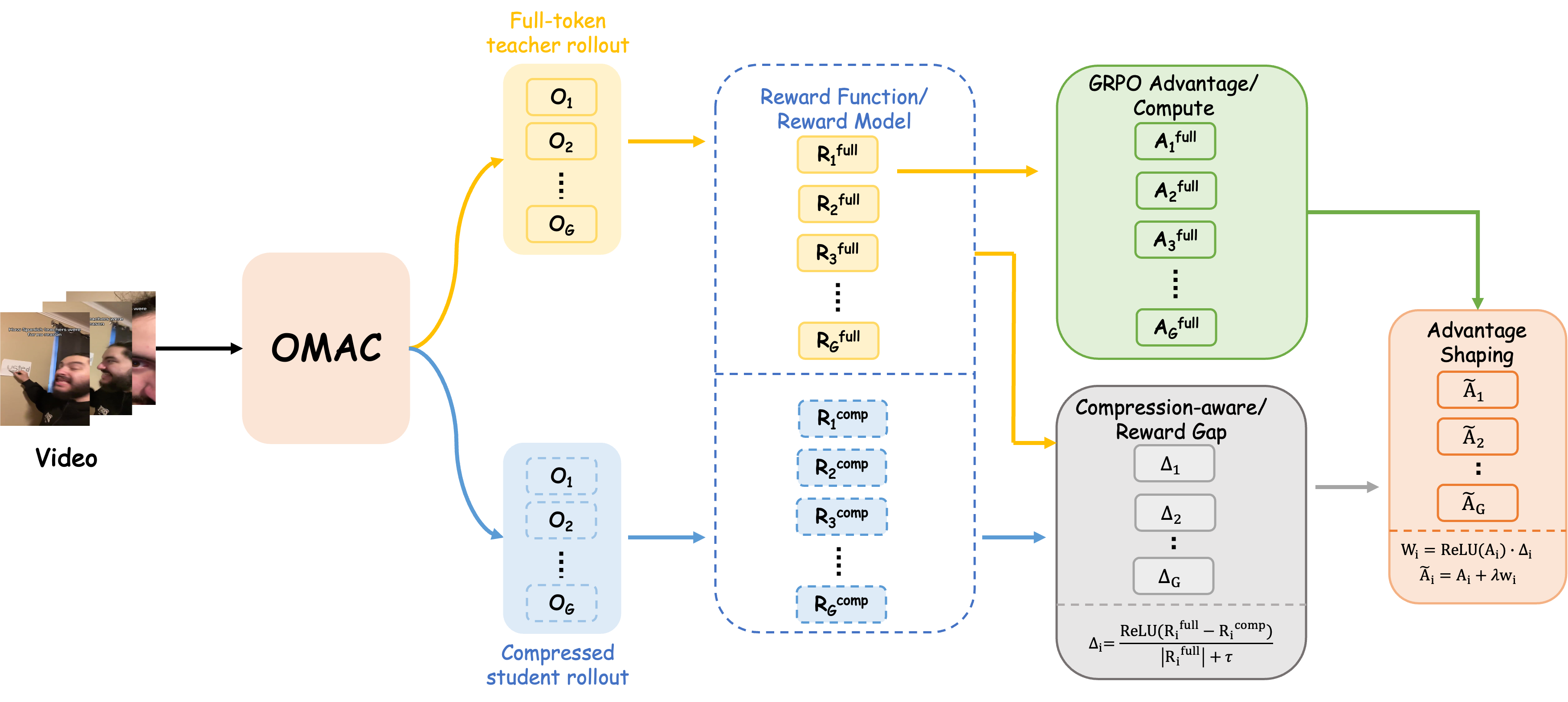}
\caption{\textbf{O-MARC for training-based compression.} The full token branch and compressed branch are sampled from the current policy, and their reward gap shapes the GRPO advantage for robust compression training.}
    \label{fig:omarc_distill}
\end{figure*}

To complement the training free OMAC compression strategy, we further introduce O-MARC, an \textit{Omni Memory-Augmented RL Token Compression} method built on GRPO~\citep{feng2026video,guo2025deepseek,wu2026st}. As shown in Figure~\ref{fig:omarc_distill} O-MARC combines memory compression with compression self-distillation: for each query $q$, we sample two rollouts from the current policy $\pi_\theta$, including a full token rollout without memory compression as the teacher branch and a compressed rollout with memory compression enabled as the student branch. Since both branches are generated online by the current policy, the entire training procedure remains on-policy.

Let $R_i^{\text{full}}$ and $R_i^{\text{comp}}$ denote the scalar rewards of the $i$-th teacher and student rollout, respectively. We use their discrepancy to measure compression-induced degradation:
\begin{equation}
\Delta_i
=
\frac{\mathrm{ReLU}(R_i^{\text{full}} - R_i^{\text{comp}})}
{|R_i^{\text{full}}| + \tau},
\end{equation}
where $\mathrm{ReLU}(\cdot)$ keeps only positive degradation, and $\tau$ is a small constant for numerical stability. A larger $\Delta_i$ indicates that compression causes a larger reward drop on the same query.

We then reshape the GRPO advantage using the teacher student reward gap. Let $A_i$ denote the original GRPO advantage of the $i$-th rollout under the standard GRPO pipeline. We define a compression-aware distillation weight
\begin{equation}
w_i = \mathrm{ReLU}(A_i)\cdot \Delta_i,
\end{equation}
where $w_i$ becomes large only when the sample is both beneficial under GRPO and strongly harmed by compression. We then use it to construct the shaped advantage
\begin{equation}
\tilde{A}_i = A_i + \lambda w_i,
\end{equation}
where $\tilde{A}_i$ is the final advantage used for policy optimization, and $\lambda$ is a hyperparameter controlling the shaping strength. Therefore, samples that are already beneficial under GRPO and are strongly degraded by compression receive larger positive updates.

We denote the clipped policy ratio by
\begin{equation}
\rho_i(\theta)=
\mathrm{clip}\!\left(
\frac{\pi_\theta(o_i \mid q)}
{\pi_{\theta_{\mathrm{old}}}(o_i \mid q)},
\; 1-\epsilon,\; 1+\epsilon
\right).
\end{equation}
Here, $\pi_{\theta_{\mathrm{old}}}$ denotes the rollout policy used to generate the current batch, and $\epsilon$ is the clipping threshold.

Finally, we replace the original advantage in the C-GRPO objective with $\tilde{A}_i$:
\begin{equation}
\small
\mathcal{L}_{\text{C-GRPO}} = 
\mathbb{E}\!\left[
\frac{1}{G}\sum_{i=1}^G \rho_i(\theta)\tilde{A}_i
- \beta\, \mathrm{KL}\bigl(\pi_\theta \,\|\, \pi_{\mathrm{ref}}\bigr)
\right].
\label{eq:cgrpo_loss}
\end{equation}
Relative to standard GRPO, our only modification is to replace the original advantage $A_i$ with the compression-aware shaped advantage $\tilde{A}_i$.

In this way, the full-token teacher branch does not supervise the compressed branch through token-level probability matching; instead, it distills its preference through advantage shaping, explicitly biasing policy optimization toward behaviors that remain robust under memory compression.

\section{Experiment}

\subsection{Experimental Settings}

\noindent\textbf{Inference Setting.} For all video experiments, we sample frames at 1 FPS and use at most 32 frames per video. The maximum single frame size is 50,174. For the audio stream, we use a sample rate of \texttt{SAMPLE\_RATE=16000}. Unless otherwise specified, compression experiments keep the retained ratio at 30\% and evaluate the resulting compressed multimodal context on downstream QA benchmarks. All results were run twice and averaged.

\noindent\textbf{Training Setting.} For O-MARC distillation, we use Qwen2.5-Omni-3B as the base model and train on 4 NVIDIA RTX PRO 6000 GPUs with 96GB memory each. The training videos follow the same preprocessing protocol as evaluation: 1 FPS sampling, at most 32 frames per video, and a maximum single frame size of 50,174. We train with \texttt{num\_train\_epochs=1}, \texttt{batch\_size=1}, learning rate $1\times10^{-5}$, and \texttt{num\_generations=4}. This 3B base model is aligned with our goal of building an efficient omnimodal model: Qwen2.5-Omni-7B and Qwen3-Omni-30B-A3B provide useful reference points, but their model scales are substantially larger and therefore less consistent with our efficiency oriented setting. For more training details you can see in appendix~\ref{sec:appendix_training_details}.

\subsection{Benchmarks}

\begin{table}[t!]
    \captionsetup{type=table}
    \centering
    \fontsize{6.5pt}{7pt}\selectfont
    \setlength\tabcolsep{3pt}
    \renewcommand{\arraystretch}{1.15}

    \begin{adjustbox}{width=\columnwidth}
    \begin{tabular}{r|cc|cccc}
    & & &
    \rotatebox{75}{AVEP} &
    \rotatebox{75}{AVST} &
    \rotatebox{75}{CSAVA} &
    \rotatebox{75}{FGAVC} \\
    Methods & Rank & Avg. &
    \multicolumn{4}{c}{\cellcolor{cyan!10}UGC-AVQA} \\
    \hline
    \multicolumn{1}{l|}{\textcolor{black}{\textit{Proprietary Models (API)}}} & & & & & & \\
    Gemini-3.1-pro & 1 &  69.1 & 68.7 & 71.4  & 68.5 & 68.0  \\
    Gemini-3.1-flash-lite & 2 &  67.1 & 68.5 & 68.0  & 64.3  & 67.5 \\

    \hline
    \multicolumn{1}{l|}{\textcolor{black}{\textit{Open-source Models}}} & & & & & & \\
    Gemma-4-E4B-it & 4 &  59.7 & 58.5 & 61.7  & 55.6 & 63.1 \\
    Gemma-4-E2B-it & 9 &  45.6 &  46.1 & 45.2 & 42.0 & 49.0 \\
    Gemma-3n-E4B-it & 5 &  54.6& 58.0 & 57.0  & 47.6 & 55.6 \\

    Gemma-3n-E2B-it & 7 & 53.0 & 55.8 & 55.6 & 46.4 & 54.4 \\

    Qwen2.5-Omni-7B & 6 & 54.2 & 52.7 & 54.4 & 51.2 & 58.7 \\

    Qwen2.5-Omni-3B & 8 & 49.0 & 44.7 & 52.4 & 48.3 & 50.5 \\

    Qwen3-Omni-30B-A3B & 3 & 65.0 & 65.3 & 66.8 & 60.9 & 67.0\\

    \end{tabular}
    \end{adjustbox}

    \vspace{-0.2cm}
    \caption{\textbf{Benchmark results on UGC-AVQA.} We report average accuracy and category accuracy for proprietary and open source omnimodal models. AVEP, AVST, CSAVA, and FGAVC denote event progression, scene or temporal reasoning, cross scene alignment, and fine grained contrast.}
    \label{tab:main_table}
    \vspace{-0.4cm}
\end{table}

The benchmark study first asks whether UGC-AVQA exposes a meaningful gap in audio visual association. Proprietary models remain strongest, with Gemini-3.1-pro and Gemini-3.1-flash-lite reaching 69.1 and 67.1 average accuracy. Among open source models, the much larger Qwen3-Omni-30B-A3B reaches 65.0, while Qwen2.5-Omni-3B only reaches 49.0. This gap supports our dataset motivation: reliable UGC audio visual QA requires more than generic visual recognition or unimodal priors.

The category breakdown further shows where models struggle. CSAVA is the lowest category for several open source models, including Qwen3-Omni-30B-A3B at 60.9 and Gemma-3n-E4B-it at 47.6. This suggests that cross scene audio visual alignment is a core difficulty of UGC-AVQA, and motivates compression methods that preserve temporally aligned acoustic and visual evidence rather than only reducing visual redundancy.

\begin{table*}[t!]
\centering
\captionsetup{skip=4pt}
\renewcommand{\arraystretch}{0.94}
\setlength{\tabcolsep}{6pt}
\resizebox{1\linewidth}{!}{
\begin{tabular}{cccccccccccc}
\toprule
\multirow{2}{*}{Method} & \multicolumn{2}{c}{Settings} & DailyOmni & \multicolumn{5}{c}{UGC-AVQA} & OmniVideo & WorldSense & Average \\
\cmidrule(lr){2-3} \cmidrule(lr){4-4} \cmidrule(lr){5-9} \cmidrule(lr){10-10} \cmidrule(lr){11-11} \cmidrule(lr){12-12}
& Retained Ratio & Frames
& Overall Accuracy
& AVEP
& AVST
& CSAVA
& FGAVC
& Overall Accuracy
& Avg. Score
& Overall Accuracy
& ACC \\
\midrule
\multicolumn{12}{c}{\emph{Qwen3-Omni-30B-A3B}} \\
\midrule
Full Tokens      & 100\% & 32 & 69.9 & 65.3 & 66.8 & 60.9 & 67.0 & 65.0 & 39.4 & 52.5 & 56.7 \\
OmniZip          & 30\%  & 32 & 61.1 & 51.5 & 56.6 & 54.4 & 53.4 & 53.9 & 31.7 & 47.4 & 48.5 \\
OMAC (Ours)      & 30\%  & 32 & \textbf{63.2} & 57.3 & 59.0 & 53.9 & 57.5 & \textbf{56.9} & \textbf{35.5} & \textbf{48.0} & \textbf{50.9} \\
\midrule
\multicolumn{12}{c}{\emph{Qwen2.5-Omni-7B}} \\
\midrule
Full Tokens      & 100\% & 32 & 56.3 & 52.7 & 54.4 & 51.2 & 58.3 & 54.2 & 34.6 & 43.6 & 47.2 \\
OmniZip          & 30\%  & 32 & 51.8 & 51.5 & 53.6 & 50.0 & 55.8 & 52.7 & 30.0 & 39.6 & 43.5 \\
OMAC (Ours)      & 30\%  & 32 & 53.6 & 52.4 & 54.9 & 49.3 & 56.3 & 53.2 & 30.9 & 42.4 & 45.0 \\
OmniSIFT\(^\textbf{*}\) (100k+ train) & 30\% & 32 & 59.0 & 52.9 & 55.8 & 55.1 & 61.7 & 56.4 & 35.0 & \textbf{44.4} & 48.7 \\
O-MARC (3k+ train) & 30\% & 32 & 54.6 & 56.3 & 64.1 & 51.2 & 65.3 & 59.2 & 31.5 & 43.4 & 47.2 \\
O-MARC\(^\textbf{*}\) (100k+ train) & 30\% & 32 & \textbf{60.4} & 65.7 & 68.2 & 58.3 & 66.3 & \textbf{64.6} & \textbf{35.2} & 44.0 & \textbf{51.1} \\
\midrule
\multicolumn{12}{c}{\emph{Qwen2.5-Omni-3B}} \\
\midrule
Full Tokens      & 100\% & 32 & 53.8 & 44.7 & 52.4 & 48.3 & 50.5 & 49.0 & 31.6 & 42.2 & 44.1 \\
OmniZip          & 30\%  & 32 & 47.7 & 44.7 & 49.0 & 44.9 & 49.8 & 47.1 & 29.3 & 39.7 & 41.0 \\
OMAC (Ours)      & 30\%  & 32 & 49.9 & 47.6 & 52.4 & 47.1 & 49.5 & 49.2 & 30.8 & 41.2 & 42.8 \\
O-MARC (3k+ train) & 30\% & 32 & \textbf{52.3} & 52.4 & 61.4 & 55.8 & 61.9 & \textbf{57.9} & \textbf{31.0} & \textbf{42.1} & \textbf{45.8} \\
\bottomrule
\end{tabular}
}
\caption{\textbf{Main comparison of token compression methods.} We compare full token inference, OmniZip, OMAC, O-MARC, and the contemporary OmniSIFT baseline under the same frame budget. DailyOmni and WorldSense report accuracy, OmniVideo reports average score, and UGC-AVQA reports both category and overall accuracy. Average ACC is computed over the main scores of DailyOmni, UGC-AVQA, OmniVideo, and WorldSense. \textit{Note(\textbf{*}): Omnisift is a concurrent project that was open-sourced in May.}}
\label{tab:main-exp}
\vspace{-3mm}
\end{table*}

\subsection{Main Results of OMAC and O-MARC}

The main comparison evaluates two claims: whether OMAC is a stronger plug in compressor than direct token pruning, and whether O-MARC can train models to reason better under compressed memory. For training free compression, OMAC consistently improves over OmniZip~\citep{tao2025omnizip}. On Qwen3-Omni-30B-A3B, the average score increases from 48.5 to 50.9, with UGC-AVQA rising from 53.9 to 56.9 and OmniVideo from 31.7 to 35.5. On Qwen2.5-Omni-7B, OMAC improves the average score from 43.5 to 45.0 and raises WorldSense from 39.6 to 42.4. These gains indicate that preserving visual memory and audio anchors keeps more reasoning evidence than direct pruning at the same retained ratio.

The larger improvement comes from compression aware training. With only 3k training samples, O-MARC on Qwen2.5-Omni-3B improves the average score from 42.8 with OMAC to 45.8 and surpasses the full token 3B baseline. On UGC-AVQA, it raises accuracy from 49.2 to 57.9, showing that the compact model learns to use compressed multimodal memory rather than merely tolerate it. Scaling the same training framework to Qwen2.5-Omni-7B further improves the average score from 45.0 to 47.2 with 3k samples, and to 51.1 with 100k samples.

We also compare with OmniSIFT~\citep{ding2026omnisift}, a contemporary compression training work released in May. OmniSIFT uses 107k training samples; for a comparable data scale, our 100k plus setting first uses 104k samples for SFT alignment and then applies RL on 3k UGC samples. Under the same 7B backbone, O-MARC achieves a higher average score than OmniSIFT, 51.1 versus 48.7, and substantially higher UGC-AVQA accuracy, 64.6 versus 56.4. OmniSIFT is slightly stronger on WorldSense, but O-MARC is stronger on DailyOmni, UGC-AVQA, and OmniVideo. This comparison suggests that explicitly training with memory compressed audio visual evidence is especially beneficial for tasks requiring fine grained multimodal association.

\subsection{Ablation Studies}

\begin{table}[t!]
    \centering
    \captionsetup{type=table}
    \fontsize{8pt}{9pt}\selectfont
    \setlength{\tabcolsep}{8pt}
    \renewcommand{\arraystretch}{1.15}

    \begin{tabular}{lcc}
    \toprule
    Method & Compression Ratio & Avg. \\
    \midrule
    Baseline & 0\% & 49.0 \\
    OmniZip & 55\% & 48.6 \\
    OmniZip & 70\% & 47.1 \\
    OMAC & 55\% & 49.5 \\
    OMAC & 70\% & 49.2 \\
    \bottomrule
    \end{tabular}

    \vspace{-0.2cm}
    \caption{\textbf{Effect of compression ratio on UGC-AVQA.} OMAC is compared with OmniZip and the full token baseline under moderate and strong compression.}
    \label{tab:ablation_ratio_ugc}
    \vspace{-0.4cm}
\end{table}

\noindent \textbf{Compression Ratio}. This ablation tests whether OMAC remains useful as the token budget changes. At 55\% compression, OMAC reaches 49.5 average accuracy, above both full token inference at 49.0 and OmniZip at 48.6. At 70\% compression, OMAC still reaches 49.2, while OmniZip drops to 47.1. The pattern supports the role of memory tokens: they are most valuable when the budget is limited and the model must keep only task relevant evidence.

\begin{table}[t!]
    \centering
    \captionsetup{type=table}
    \fontsize{8pt}{9pt}\selectfont
    \setlength{\tabcolsep}{8pt}
    \renewcommand{\arraystretch}{1.15}

    \begin{tabular}{lcccc}
    \toprule
    Method & Frames & Size & Avg.  \\
    \midrule
    Baseline & 64 & \(128\times28\times28\) & 43.6 \\
    Baseline & 32 & \(64\times28\times28\) & 42.2 \\
    OmniZip & 64 & \(128\times28\times28\) & 40.2 \\
    OmniZip & 32 & \(64\times28\times28\) & 39.7 \\
    OMAC & 64 & \(128\times28\times28\) & 41.7 \\
    OMAC & 32 & \(64\times28\times28\) & 41.2 \\
    \bottomrule
    \end{tabular}

    \vspace{-0.2cm}
    \caption{\textbf{Effect of video size on WorldSense.} We compare OMAC and OmniZip with different frame counts and maximum visual token sizes on long video examples.}
    \label{tab:ablation_worldsense_size}
    \vspace{-0.4cm}
\end{table}

\noindent \textbf{Video Size}. The video size ablation asks whether OMAC is robust when the visual input is shortened. On WorldSense, OMAC improves over OmniZip from 40.2 to 41.7 with 64 frames, and from 39.7 to 41.2 with 32 frames. The gain remains stable even under the smaller visual budget, indicating that OMAC does not simply benefit from more frames, but from allocating the retained memory more effectively.

\begin{table}[t!]
    \centering
    \captionsetup{
        type=table
    }
    \fontsize{8pt}{9pt}\selectfont
    \setlength{\tabcolsep}{6pt}
    \renewcommand{\arraystretch}{1.15}

    \centering
    \begin{tabular}{lcccc}
    \toprule
    Method & Frames & Max Size & Time(s) & Memory(GB) \\
    \midrule

    Baseline & 64  & \(128\times28\times28\) & 3.81 & 24.2 \\
    OmniZip  & 64  & \(128\times28\times28\) & 2.47 & 15.7 \\
    OMAC   & 64  & \(128\times28\times28\) & 2.49 & 15.8 \\
    \bottomrule
    \end{tabular}
    \vspace{-0.1cm}
    \caption{\textbf{Inference cost on long video.} We report latency and GPU memory on WorldSense.}
    \label{tab:latency}
    \vspace{-0.35cm}
\end{table}

\noindent \textbf{Efficiency}. The cost experiment verifies that the accuracy gains do not come from a heavier inference path. Compared with full token inference, OMAC reduces latency from 3.81s to 2.49s and memory from 24.2GB to 15.8GB. Its cost is almost the same as OmniZip, which uses 2.47s and 15.7GB, but OMAC gives stronger accuracy in the main comparison and ablations. Thus, OMAC keeps the efficiency profile of token compression while preserving more reasoning evidence.

\begin{table}[t!]
    \centering
    \captionsetup{
        type=table
    }
    \fontsize{8pt}{9pt}\selectfont
    \setlength{\tabcolsep}{6pt}
    \renewcommand{\arraystretch}{1.15}

    \centering
    \begin{tabular}{lccc}
    \toprule
    Method & Frames & Joint Way & DailyOmni \\
    \midrule

    Baseline & 32  & NA & 53.8 \\
    OmniZip  & 32  & A-V & 47.7 \\
    OMAC(*) & 32  & V-A & 48.2 \\
    OMAC   & 32  & V-A-VA & 49.9 \\
    \bottomrule
    \end{tabular}
    \vspace{-0.1cm}
\caption{\textbf{Guidance and intervention phase influence.} A-V denotes audio guided visual compression, V-A denotes visual guided audio compression, and V-A-VA denotes the full OMAC design that first builds visual and audio memory and then uses visual memory to adjust the audio budget.}
    \label{tab:guidance_phase}
    \vspace{-0.35cm}
\end{table}

\noindent \textbf{Guidance Phase}. This design ablation studies how audio and visual compression should interact. Audio guided visual compression, as in OmniZip, reaches 47.7 on DailyOmni. Directly using visual information to guide audio compression improves the score to 48.2, showing that visual context is useful for locating important acoustic regions. The full OMAC design reaches 49.9 by first constructing visual and audio memory separately, then using the retained visual memory to adjust the audio budget. This supports our final design choice: the audio budget should be guided by visual memory after salient visual evidence has been selected, rather than by raw visual signals alone.

\section{Conclusion}

We present a unified study of efficient omnimodal reasoning through benchmark construction, memory compression, and compression aware training. UGC-AVQA targets audiovisual association in public UGC videos and uses audio removal filtering to ensure that evaluation requires both acoustic and visual evidence. OMAC introduces a training free memory compression mechanism that preserves salient visual memory and temporally grounded audio anchors at nearly the same inference cost as direct pruning. Building on this compressed memory, O-MARC trains compact models to reason under a reduced token budget. Experiments show that OMAC consistently improves over OmniZip across model scales, while O-MARC achieves the strongest gains in the efficient 3B setting and outperforms the contemporary OmniSIFT when trained with comparable 100k plus data. These results indicate that audiovisual compression should be viewed not only as an inference acceleration technique, but also as a trainable condition for robust and efficient omnimodal reasoning.

\section*{Limitations}

UGC-AVQA is built from publicly accessible short video platforms, so its coverage naturally reflects the style, language, and content distribution of currently available UGC videos. Since we release video links rather than raw files, a small portion of the data may become unavailable over time as platform content changes. Future work can extend the benchmark to more languages, longer videos, and additional real world domains while keeping the same audio visual dependency focused evaluation protocol.




\bibliography{custom}

\appendix

\section{Appendix}
\label{sec:appendix}

\subsection{Training Details}
\label{sec:appendix_training_details}

Our GRPO training uses four sampled rollouts per prompt with a maximum completion length of 1024 and a maximum context length of 32768. We optimize for 2 epochs with learning rate $1\times 10^{-5}$, batch size 1 per device, and 4-way generation under bfloat16 and DeepSpeed ZeRO-2. The reward is the sum of two equally weighted terms, namely answer accuracy and format correctness. For multimodal preprocessing, we use 1 FPS video sampling, cap the input to at most 32 frames, disable audio output generation, and enable audio-in-video reasoning.

\begin{figure*}[t]
\centering
\begin{subfigure}[t]{0.48\textwidth}
    \centering
    \includegraphics[width=\linewidth]{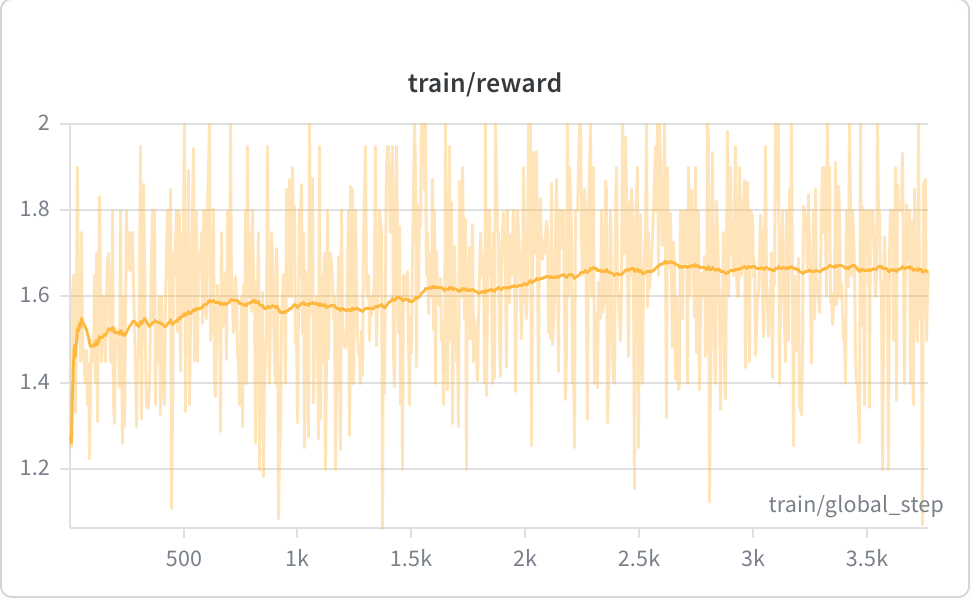}
    \caption{Training reward.}
    \label{fig:appendix_train_reward}
\end{subfigure}
\hfill
\begin{subfigure}[t]{0.48\textwidth}
    \centering
    \includegraphics[width=\linewidth]{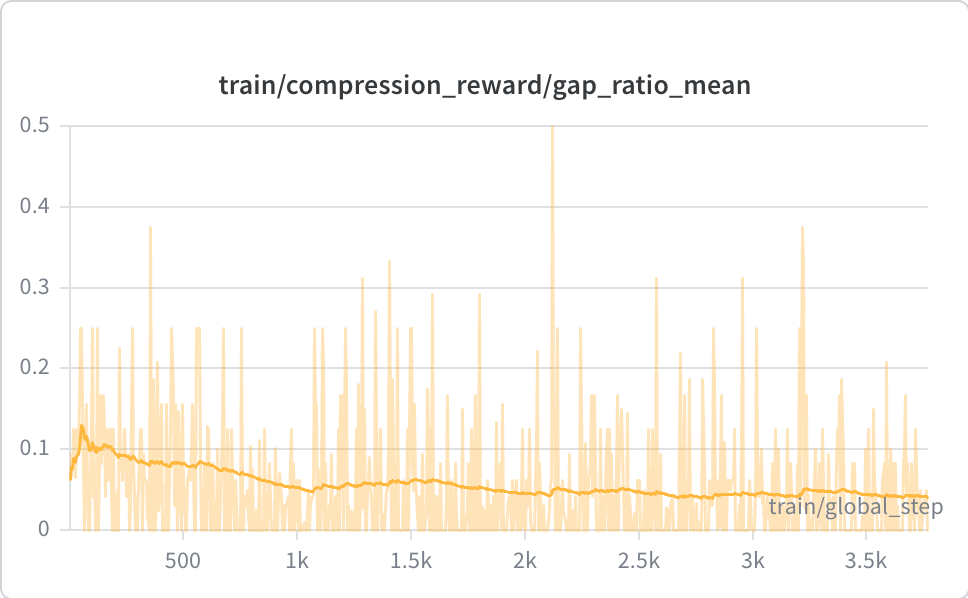}
    \caption{Compression reward gap ratio.}
    \label{fig:appendix_gap_ratio}
\end{subfigure}
\caption{\textbf{Training dynamics of O-MARC.} The total reward rises steadily during GRPO training, while the compression reward gap ratio gradually decreases, indicating that the compressed rollout becomes better aligned with the full-token teacher rollout over time.}
\label{fig:appendix_training_curves}
\end{figure*}

As shown in Figure~\ref{fig:appendix_training_curves}, the overall reward improves steadily throughout training, while the compression gap ratio exhibits a downward trend despite local fluctuations. This pattern is consistent with the goal of O-MARC: the policy becomes stronger overall and, at the same time, more robust under memory compression.

\subsection{Annotator Instructions}
\label{sec:appendix_annotator_instructions}

Human annotators were recruited from the authors' institution as student research assistants with experience in multimodal annotation, and were compensated according to local institutional standards for research assistance. Annotators were instructed that the videos and derived annotations would be used only for research on audiovisual reasoning. Annotators were asked to inspect both the visual stream and the audio track, write an omni detail caption covering visible entities, actions, scene transitions, speech, non-speech sounds, and temporal relations, and avoid adding private identity information beyond what is already visible or audible in the public video. For each video, annotators then wrote multiple choice questions whose answers require both modalities. The four target categories were defined as follows: \textbf{Audio-Visual Event Progression} asks what audio event follows or precedes a visual event; \textbf{Audio-Visual Scene or Temporal Transition} asks how audio changes across visual scene changes; \textbf{Cross-Scene Audio-Visual Alignment} asks which audio segment corresponds to a visual moment across scenes; and \textbf{Fine-Grained Audio-Visual Contrast} asks how subtle acoustic cues differ across visual states. Annotators were instructed to provide one correct option, plausible distractors, and evidence from the caption for each answer. During quality control, reviewers removed questions with ambiguous answers, visually solvable questions, malformed options, or answers unsupported by the video.

\subsection{Question Taxonomy Comparison}
\label{sec:appendix_taxonomy}

We further compare the question taxonomy of UGC-AVQA with the annotation files used by existing audiovisual benchmarks. UGC-AVQA is organized around four association centric categories: Audio-Visual Event Progression, Audio-Visual Scene or Temporal Transition, Cross-Scene Audio-Visual Alignment, and Fine-Grained Audio-Visual Contrast. These categories are designed to test whether a model can jointly inspect what is heard and what is seen, rather than only localize a relevant visual span or recognize an isolated audio event.

\begin{table*}[t]
\centering
\small
\begin{tabularx}{\textwidth}{p{0.18\textwidth}p{0.31\textwidth}X}
\toprule
Benchmark & Representative categories & Category focus \\
\midrule
UGC-AVQA & Event progression, scene or temporal transition, cross-scene alignment, fine-grained contrast & Audio and visual evidence must be compared, aligned, or contrasted across time. \\
WorldSense & Object counting, spatial relation, attribute recognition, event sorting, temporal localization, audio recognition & Broad world understanding across visual, audio, temporal, and semantic tasks. \\
OmniVideoBench & Counting, causal reasoning, temporal understanding, reference reasoning, fine-grained perception, sentiment analysis & General long video reasoning, often using speech as a cue to locate or disambiguate visual evidence. \\
DailyOmni & Event sequence, AV event alignment, context understanding, reasoning, inference, comparative & Daily video understanding with event order, scene context, and general audiovisual matching. \\
\bottomrule
\end{tabularx}
\caption{\textbf{Comparison of question taxonomies.} UGC-AVQA focuses on audio visual association, while existing benchmarks emphasize broader video understanding skills.}
\label{tab:appendix_taxonomy_comparison}
\end{table*}

Table~\ref{tab:appendix_taxonomy_comparison} shows that existing benchmarks provide valuable coverage of general video understanding, but their categories are not primarily defined by cross-modal dependence. WorldSense includes audio recognition and audio counting, while many categories still focus on visual attributes, spatial relations, or object counts. OmniVideoBench contains multimodal reasoning steps, yet the taxonomy is organized around general abilities such as counting, causal reasoning, and temporal understanding. DailyOmni includes AV Event Alignment, but it is one part of a broader daily video taxonomy. In contrast, UGC-AVQA makes audio visual association the central evaluation target.
\begin{figure*}[t]
    \centering
        \includegraphics[width=\textwidth]{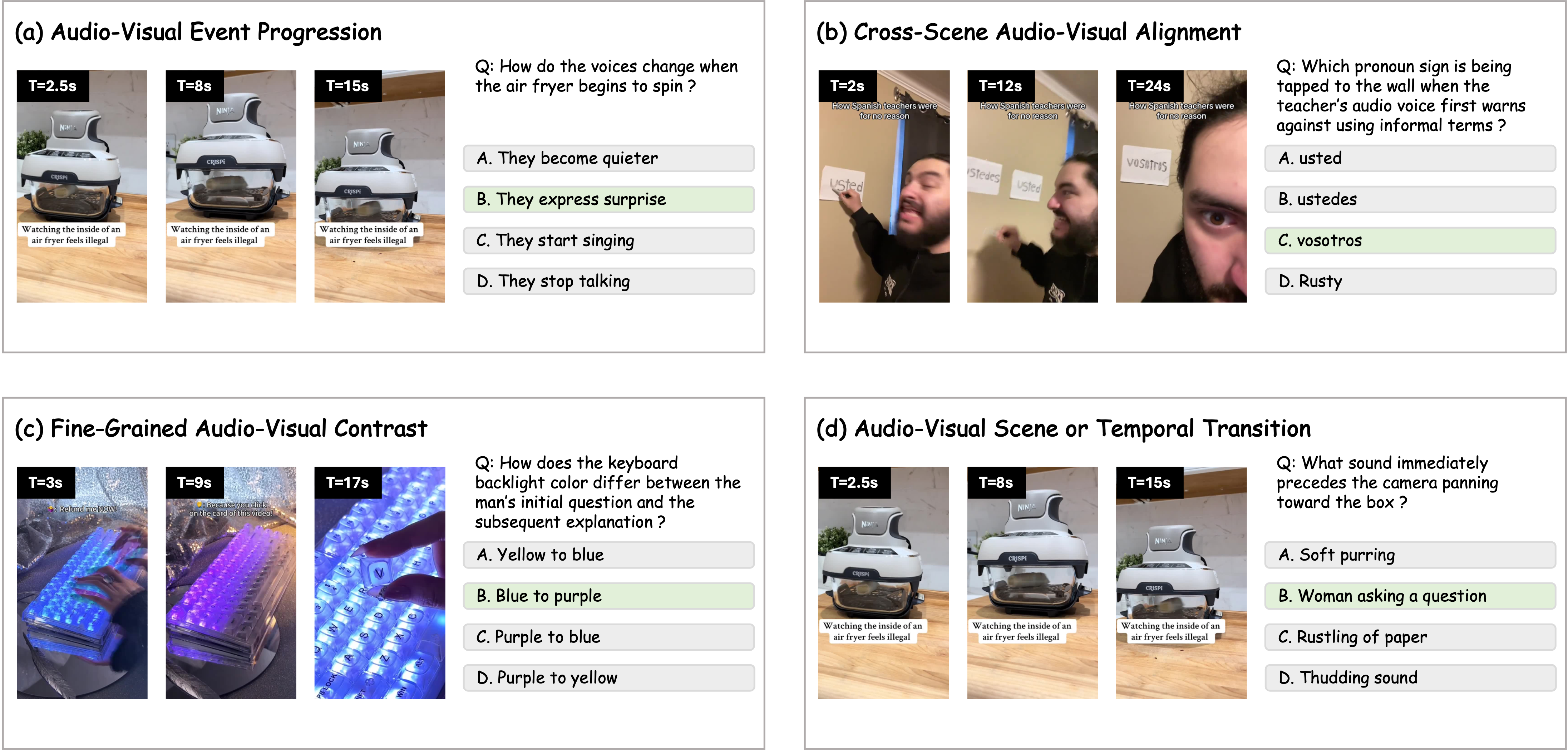}
\caption{\textbf{Representative UGC-AVQA cases.} The figure will show one example from each of the four UGC-AVQA categories: audio visual event progression, scene or temporal transition, cross-scene audio visual alignment, and fine-grained audio visual contrast.}
\label{fig:ugc_avqa_cases}
\end{figure*}

\subsection{Examples and Necessity}
\label{sec:appendix_examples}

The distinction is also visible in question examples. A WorldSense question may ask the position of an object relative to a person, which mainly tests spatial perception. An OmniVideoBench question may ask how many objects appear when a narration segment is mentioned, where audio helps locate the moment and vision provides the answer. A DailyOmni question may ask which audio event is synchronized with a visual shot, testing local alignment. UGC-AVQA instead asks questions such as what happens in the audio after a specific visual action, what audio accompanies a visual scene, or how the vocal tone changes between two visual moments. Such questions require the model to maintain both streams and reason over their relation.

This focus makes UGC-AVQA necessary for evaluating efficient omnimodal models. Compression methods may preserve visually salient frames while discarding brief audio cues, or keep audio anchors without enough visual context to interpret them. A benchmark centered on cross-modal dependency can reveal these failures more directly than broad video QA. The audio removal filtering used in UGC-AVQA further strengthens this property: a question is retained only when the answer cannot be reliably recovered from the muted video, making the benchmark a targeted test of audio visual association rather than visual prior matching.

Figure~\ref{fig:ugc_avqa_cases} illustrates the four types of association tested by UGC-AVQA. Together, these cases highlight why the dataset is important for studying audiovisual compression: a model must preserve the right acoustic cue, the right visual moment, and their temporal relation under a reduced token budget.

\end{document}